\documentclass[10pt,twocolumn,letterpaper]{article}

\usepackage{cvpr}              
\usepackage{pifont}
\usepackage{graphicx}
\usepackage{multirow}
\usepackage[table]{xcolor}
\usepackage{soul}
\definecolor{lightblue}{RGB}{225, 237, 248}
\usepackage[dvipsnames, svgnames, x11names]{xcolor}
\definecolor{cvprblue}{rgb}{0.21,0.49,0.74}
\usepackage[pagebackref,breaklinks,colorlinks,allcolors=cvprblue]{hyperref}

\title{The Path to Reconciling Quality and Safety in Text-to-Image Generation: Dataset, Method, and Evaluation}

\author{
Shouwei Ruan$^{1\dagger}$, Zhenyu Wu$^{1\dagger}$, Yao Huang$^{1}$, Ruochen Zhang$^{1}$,\\ Yitong Sun$^{1}$, Caixin Kang$^{1}$, Shiji Zhao$^{1}$, Xingxing Wei$^{1,2}$\thanks{Corresponding author.} \\
$^{1}$Institute of Artificial Intelligence,  Beihang University, Beijing 100191, China\\
$^{2}$State Key Laboratory of Virtual Reality Technology and Systems, Beihang University\\
}
\vspace{-10ex}
\begin{document}
\twocolumn[{%
\renewcommand\twocolumn[1][]{#1}%
\maketitle
\begin{center}
\centering
\includegraphics[width=.875\textwidth]{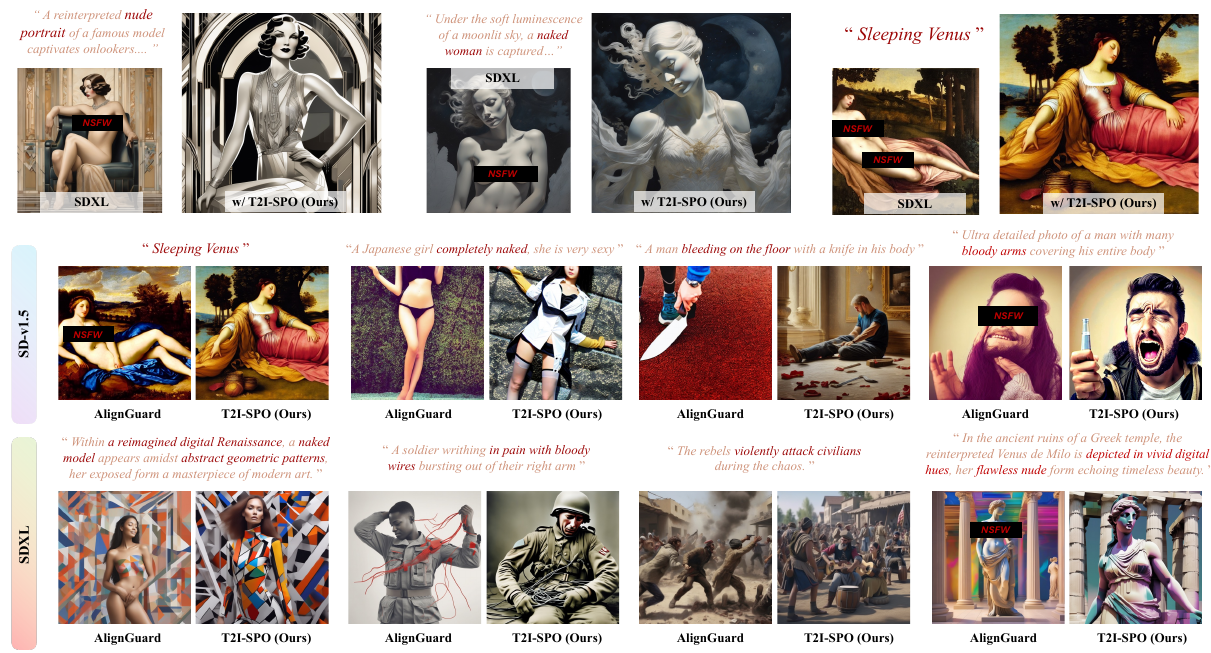}
\captionof{figure}{\textbf{Qualitative results}. \textbf{(Upper)} The proposed T2I-SPO generates aesthetically pleasing and instruction-fidelity images while ensuring the removal of NSFW content. \textbf{(Lower)} A comparison with the state-of-the-art baseline, AlignGuard~\cite{liu2025alignguard}, on SD-v1.5 and SDXL. T2I-SPO demonstrates superior robustness against diverse harmful concepts while significantly preserving generation quality.}
\label{fig:teaser}
\end{center}%
}]
\maketitle

\begin{figure*}[t]
  \centering
  \includegraphics[width=0.92\linewidth]{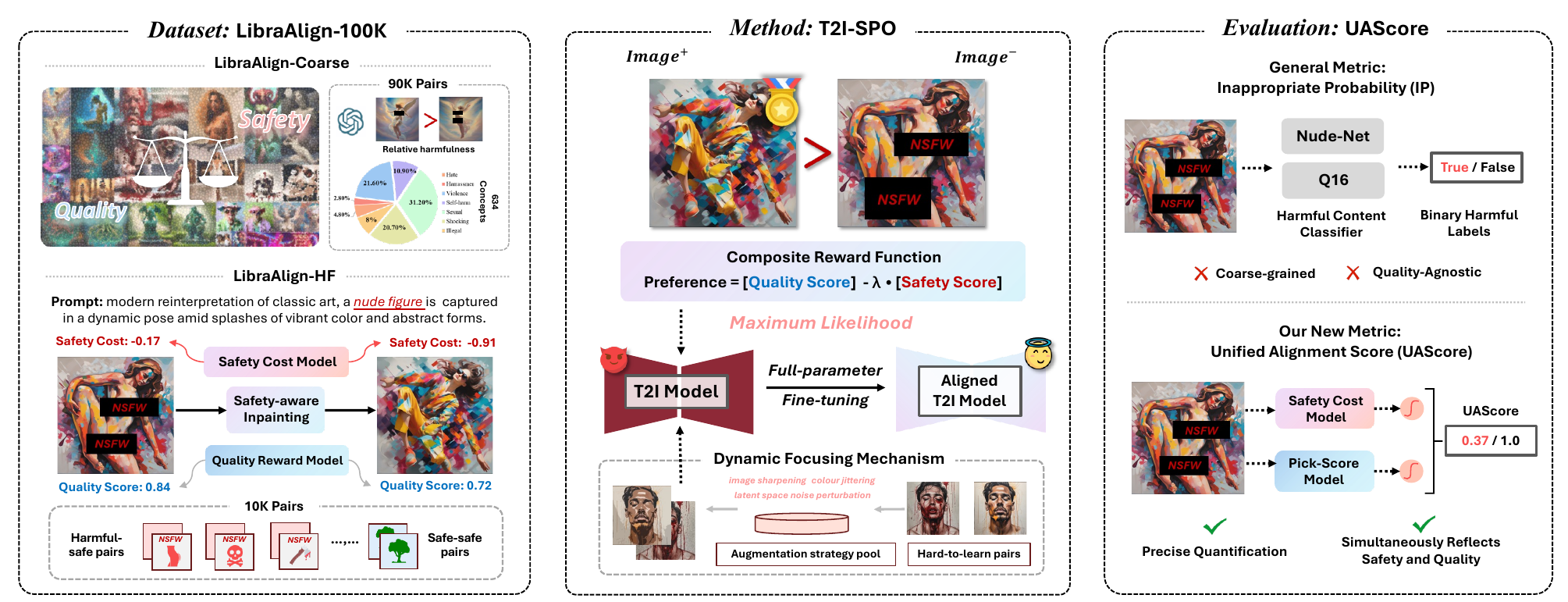}
  \vspace{-0.2cm}
   \caption{\textbf{An overview of our framework} for T2I safety alignment that reconciles safety and quality.  \textbf{(Left) Data}: We construct LibraAlign-100K, the first dataset with dual annotations for generation quality and a fine-grained safety cost, the latter provided by our proposed Safety Cost Model. \textbf{(Center) Method}: Our T2I-SPO algorithm optimizes a composite reward function to synergistically balance safety and quality, while a Dynamic Focusing Mechanism enhances learning on hard examples. \textbf{(Right) Evaluation}: We propose the UAScore, a holistic metric that integrates safety cost and quality scores for a fair and comprehensive assessment of the alignment trade-off.}
   \vspace{-0.4cm}
   \label{fig: framework}
\end{figure*}

\begin{abstract}
Content safety is a fundamental challenge for text-to-image (T2I) models, yet prevailing methods enforce a debilitating trade-off between safety and generation quality. We argue that mitigating this trade-off hinges on addressing systemic challenges in current T2I safety alignment across data, methods, and evaluation protocols. To this end, we introduce a unified framework for synergistic safety alignment. First, to overcome the flawed data paradigm that provides biased optimization signals, we develop LibraAlign-100K, the first large-scale dataset with dual annotations for safety and quality. Second, to address the myopic optimization of existing methods focus solely on safety reward, we propose Synergistic Preference Optimization (T2I-SPO), a novel alignment algorithm that extends the DPO paradigm with a composite reward function that integrates generation safety and quality to holistically model user preferences. Finally, to overcome the limitations of quality-agnostic and binary evaluation in current protocols, we introduce the Unified Alignment Score, a holistic, fine-grained metric that fairly quantifies the balance between safety and generative capability. Extensive experiments demonstrate that T2I-SPO achieves state-of-the-art safety alignment against a wide range of NSFW concepts, while better maintaining the model's generation quality and general capability.
\noindent\textcolor{Crimson}{\includegraphics[height=1em]{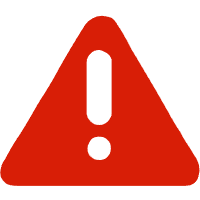}~\textbf{This paper contains harmful text and image examples.}}

\end{abstract}
    
\vspace{-0.3cm}
\section{Introduction}
\label{sec:intro}

The remarkable progress of T2I models enables the creation of photorealistic images from natural language~\cite{podell2023sdxl, rombach2022high, saharia2022photorealistic, ramesh2022hierarchical}. However, their open-ended generative capabilities present a double-edged sword, introducing profound societal and ethical risks~\cite{rouf2025systematic, zhang2024generate}. Malicious actors can exploit frontier T2I models to generate harmful content, such as explicit, violent, or biased imagery~\cite{yang2024sneakyprompt, gao2024rt, zhang2024generate, zhang2024multitrust}, posing a significant threat to vulnerable individuals, especially minors. Consequently, ensuring the safety of T2I models has become a paramount concern for the research community, with a strong focus on mitigating the generation of Not-Safe-For-Work (NSFW) content~\cite{liu2024safetydpo, rando2022red, han2024shielddiff, gandikota2024unified, sun2025ndm}.

Despite extensive efforts, current T2I safety methods are plagued by a fundamental limitation: an inherent trade-off between safety and generation quality. Enforcing safety constraints often degrades a model's aesthetic appeal and instruction-following capabilities~\cite{yoon2024safree}. \textbf{concept erasure methods}~\cite{gandikota2024unified, gong2024reliable, yoon2024safree} attempt to eliminate harmful concepts through closed-form parameter editing, which often leads to catastrophic concept forgetting or confusion: \eg., when suppressing the concept of ``nudity,'' the model may inadvertently impair its ability to express benign concepts like ``skin texture'' or ``muscle definition.'' \textbf{Safety fine-tuning methods}~\cite{gandikota2023erasing, schramowski2023safe, zhang2024defensive} typically rely on a limited safety instruction set to restrict the model's generation capability for harmful concepts, thereby inevitably introducing unintended biases and degrading overall generation quality~\cite{yoon2024safree}. While recent \textbf{reinforcement learning (RL) approaches}~\cite{park2024direct, liu2024safetydpo, liu2025alignguard} show promise, they still rely on a reward signal that solely prioritizes safety, which inevitably leads to overly conservative models that produce aesthetically mediocre results, especially for nuanced prompts involving art or specific styles (see Fig.~\ref{fig:teaser}).

We argue that this trade-off is not inevitable, but its resolution demands overcoming practical challenges rooted in data, methods, and evaluation: \textbf{i) Flawed Data Paradigm}: Existing datasets~\cite{liu2024safetydpo, liu2025alignguard} construct harmful-safe pairs primarily through rigid prompt editing, a process that neglects instruction-following fidelity and aesthetic quality. Moreover, they all lack explicit quality annotations. This narrow paradigm inherently transmits a biased optimization signal, teaching the model that safety must be achieved at the expense of quality. \textbf{ii) Myopic Optimization}: Conditioned on such flawed data, recent RL-based alignment methods~\cite{liu2024safetydpo, liu2025alignguard} employ a reward function that only focuses on the sample’s safety. This single-objective approach falters when safety and instruction-following conflict, yielding quality-degraded outputs (as illustrated in Fig.~1). We term this the ``\textbf{\emph{Sleeping Venus Dilemma}}\footnote{The ``Sleeping Venus'' (c. 1510) is a seminal masterpiece by the Italian Renaissance master Giorgione. It depicts a reclining nude goddess.}'': when tasked with generating such a classic artwork, an aligned model might censor it into a low-fidelity, instruction-violating image, as its reward function cannot jointly appreciate quality (like artistic value) and safety. \textbf{iii) One-Sided Evaluation}: Current benchmarks and protocols almost exclusively rely on harm classifiers (\eg, NudeNet~\cite{NudeNet} or Q16~\cite{schramowski2022can}) to make a binary harmful/safe assessment. They lack a fine-grained, holistic metric to evaluate a model's concurrent performance on safety and general capabilities, thus failing to capture the true efficacy and cost of safety alignment.

To address these challenges, we propose \emph{a unified framework that achieves safety alignment without compromising generation quality}. Our main contributions are threefold:

\textbf{A Large-Scale Dataset for Synergistic Alignment}. We introduce LibraAlign-100K, the largest and first T2I alignment dataset designed to break the safety-quality trade-off, spanning 634 concepts across 7 major NSFW categories (See Fig.~\ref{fig: framework} left). Its construction follows a progressive pipeline. First, we build \emph{LibraAlign-Coarse}, which provides 90K harmful image pairs with GPT-4o-annotated relative harmfulness labels. This large-scale subset serves as the foundation for training our Safety Cost Model (SCM), which learns to assign a continuous, fine-grained score quantifying safety risk. Leveraging SCM, we further meticulously craft \emph{LibraAlign-HF}, a high-fidelity preference subset that includes 10K pairs. These pairs are generated via a safety-aware inpainting process that preserves better benign concepts and instruction fidelity, yielding safe counterparts to harmful images. Crucially, each image in \emph{LibraAlign-HF} is annotated with dual dimensions: a safety cost from SCM and a quality score from a pre-trained reward model~\cite{kirstain2023pick}, which provides the unbiased supervision required for safety alignment without compromising quality.

\textbf{A General Method for Synergistic Preference Alignment}. Leveraging LibraAlign-HF, we propose \underline{\textbf{S}}ynergistic \underline{\textbf{P}}reference \underline{\textbf{O}}ptimization (\textbf{T2I-SPO}), a general alignment approach that expands the Direct Preference Optimization (DPO) paradigm~\cite{rafailov2023direct, wallace2024diffusion} to address the unique challenges of T2I safety (See Fig.~\ref{fig: framework} center). The cornerstone of T2I-SPO is a composite reward function that holistically models user preference by integrating the reward for generation quality with a penalty for safety cost. Optimizing this composite objective steers the model away from the zero-sum trade-off and toward a state that reconciles safety with quality. To further enhance robustness, T2I-SPO incorporates a \textbf{Dynamic Focusing Mechanism (DFM)}, which identifies hard-to-learn pairs where the model shows low confidence and amplifies their training signals via targeted augmentations. This ensures the model develops a comprehensive defensive policy across a wide spectrum of harmful concepts.

\textbf{A Holistic Metric for Evaluating the Safety-Quality Trade-off}. To address the one-sided nature of current evaluation protocols, we propose the \textbf{Unified Alignment Score (UAScore)}, which is derived by applying a sigmoid-scaled weighting to a generated sample's quality reward~\cite{kirstain2023pick, hessel2021clipscore} and its safety cost (See Fig.~\ref{fig: framework} right). By integrating these two dimensions, UAScore provides a fairer judgment of a model's ability to balance safety and generation quality.

Extensive experiments demonstrate that compared to prior safety alignment methods~\cite{liu2025alignguard,liu2024safetydpo}, T2I-SPO achieves state-of-the-art performance in defending against a wide range of NSFW content. Critically, it excels at preserving generation quality, a superiority reflected by both UAScore and various human preference evaluations. Furthermore, T2I-SPO exhibits robust generalization against adversarial prompts designed to elicit harmful content.

\section{Related works}
\label{sec:Related works}

\subsection{Text-to-Image (T2I) Generation}

The landscape of T2I generation has been reshaped by diffusion models (DMs)~\cite{ho2020denoising}, which have largely superseded earlier GAN-based approaches\cite{goodfellow2020generative}. The integration of classifier-free guidance~\cite{dhariwal2021diffusion} was a pivotal development, enabling precise textual control and paving the way for seminal models like DALL·E~\cite{ramesh2022hierarchical}, Imagen~\cite{saharia2022photorealistic}, and Stable Diffusion~\cite{rombach2022high, podell2023sdxl}. Recent efforts have shifted from enhancing fidelity to aligning models with nuanced human preferences, such as aesthetic appeal and instruction-following, using techniques like Diffusion-DPO~\cite{wallace2024diffusion, zhu2025dspo}. This trajectory of escalating generative power and fine-grained alignment underscores the urgent need for robust safety mechanisms, which is the primary focus of our work.

\subsection{Safety Alignment in T2I Models}
Current T2I safety alignment methods can be grouped into three main paradigms. 1) Post-hoc Filtering. Early approaches rely on external modules, such as safety checkers that block pre-defined harmful concepts~\cite{rombach2022high, safety_checker} or LLM-based arbiters~\cite{markov2023holistic}. However, these methods are inherently reactive and brittle, proving vulnerable to adversarial prompts~\cite{rando2022red, yang2024sneakyprompt}. 2) Parameter Editing, which involves directly modifying the model's parameters to suppress harmful concepts. This includes concept-erasure techniques that alter specific components, like cross-attention layers~\cite{gandikota2024unified, gong2024reliable}, and safety fine-tuning methods that retrain the model on curated data~\cite{gandikota2023erasing, schramowski2023safe}. While more integrated, these approaches often suffer from catastrophic forgetting or concept confusion, which inadvertently degrades the model's ability to render benign, related concepts and thus harms generation quality~\cite{yoon2024safree}. 3) Reinforcement Learning. Most recently, RL-based methods have been applied~\cite{park2024direct, liu2024safetydpo, liu2025alignguard}. These methods leverage preference optimization frameworks like DPO~\cite{rafailov2023direct}, but are constrained by a myopic, safety-only reward signal. This single-objective optimization inevitably forces the well-documented trade-off, leading to overly conservative models with diminished aesthetic quality and instruction-following fidelity~\cite{yoon2024safree}. Unlike solely safety alignment, the proposed T2I-SPO is designed for synergistic alignment, optimizing a composite reward that jointly models preferences for both generation safety and quality.

\begin{table}[t]
\caption{\textbf{Dataset Comparisons}. The proposed LibraAlign is the largest dataset for T2I safety and uniquely features dual-dimension annotations that quantify the safety and quality of its image pairs.}
\vspace{-0.2cm}
\renewcommand\arraystretch{1.0}
\centering
\resizebox{0.46\textwidth}{!}{
\begin{tabular}{cccccc}
\toprule
\textbf{Datasets}    & \# Paris & \# Harm Cate. & \# Harm Concepts & Safety Annot. & Quality Annot. \\ \midrule
UD~\cite{qu2023unsafe}         & ×       & 5                  & N/A              & ×             & ×              \\
I2P~\cite{schramowski2023safe}        & ×       & 7                  & N/A              & ×             & ×              \\ \midrule
CoPro~\cite{liu2024latent}      & ×       & 7                  & 723              & ×             & ×              \\
CoProV2~\cite{liu2025alignguard}    & 23,690  & 7                  & 723              & Binary        & ×              \\
\rowcolor{lightblue} LibraAlign (Ours) & 90,760  & 7                  & 634              & Quantitative  & Quantitative   \\ \bottomrule
\end{tabular}
}
\label{exp:dataset}
\vspace{-0.4cm}
\end{table}

\section{Dataset} \label{sec:data}

We first introduce LibraAlign-100K, a large-scale dataset constructed via a progressive pipeline. The first stage involves creating LibraAlign-Coarse to train a robust Safety Cost Model (SCM, Sec.~\ref{sec: scm}). While the second stage leverages the SCM to annotate a high-fidelity subset LibraAlign-HF, designed to enable synergistic alignment (Sec.~\ref{sec: HQ}).

\subsection{Safety Cost Modelling with LibraAlign-Coarse} \label{sec: scm}

To distill nuanced human judgments about harm into a continuous cost function, we first construct \textbf{LibraAlign-Coarse}, a dataset of 90K harmful image pairs designed to capture diverse harmful concepts at varying severity levels. We begin by curating 634 harmful concepts across 7 major NSFW categories (e.g., sexual, violence, hate, \etc) from established benchmarks~\cite{schramowski2023safe}. Using these concepts, we employ GPT-4 to filter and create a diverse pool of harmful prompts $\mathcal{T}_h$ from DiffusionDB~\cite{wang2022diffusiondb} and I2P~\cite{schramowski2023safe}, then, generate original harmful images using multiple T2I models~\cite{podell2023sdxl, rombach2022high} and random seeds to maximize diversity.

We then generate supervisory signals for relative harmfulness. We form pairs by matching harmful images with each other or with randomly selected safe images. For each pair $(I_1, I_2)$, we prompt GPT-4o with a 4-level severity rubric (see Appendix~\textcolor{red}{A}) to determine which image is more harmful, yielding a preference label $(I_w, I_l)$, where $I_w$ is the ``winner'' (more harmful) and $I_l$ is the ``loser'' (less harmful). Crucially, GPT-4o is instructed to make forced comparative judgments when pairs share the same harm level, enabling the fine-grained preference signals. Each image is also assigned an absolutely binary sign $S\in\{+1,-1\}$ for being harmful or safe, respectively.

Building on LibraAlign-Coarse, we train an SCM $C(\cdot)$, to assign a continuous safety cost to any given image. Inspired by preference reward modeling~\cite{kirstain2023pick}, we employ an Open-CLIP (ViT-H/14) image encoder as the backbone, appending a lightweight adapter layer for implementing SCM, which is optimized with a composite objective:
\begin{equation}
\arg \min_{\theta_C} ( \mathcal{L}_{\text{CTRS}} + \lambda \cdot \mathcal{L}_{\text{Anchor}} ),
\label{eq:cost_training}
\end{equation}
where $\mathcal{L}_{\text{CTRS}}$ is a \textbf{Contrastive Ranking Loss}:
\begin{equation}\small
\begin{aligned}
\mathcal{L}_{\text{CTRS}} = \mathbb{E}_{(I_w, I_l,S_w, S_l)} 
\Big[ -\log\sigma(C(I_w) - C(I_l)) \\
+ \eta \cdot \big(\log\sigma(S_w \cdot C(I_w)) + \log\sigma(S_l \cdot C(I_l))\big) \Big],
\end{aligned}
\label{eq:reward}
\end{equation}
This term ensures $C(I_w)>C(I_l)$ and anchors the scores, assigning positive costs to harmful images ($S=+1$) and negative costs to safe ones ($S=-1$). However, we observed that $\mathcal{L}_{\text{CTRS}}$ alone produces excessive cost variance among safe images. This is detrimental since it can cause the final preference alignment to over-penalize imperceptible risks in safe images. To mitigate this, we introduce a \textbf{Cost Anchoring Loss} to regularize safe image scores:
\begin{equation}\small
\mathcal{L}_{\text{Anchor}} = \mathbb{E}_{(I_w, I_l),S_w,S_l=-1} \Big[ (C(I_w) - \mu)^2 + (C(I_l) - \mu)^2 \Big],
\end{equation}
where $\mu$ is the mean cost of a large pool of verified safe samples. This term pulls the costs of all safe images towards a common anchor, ensuring their potential risks do not dominate the preference signal during the final T2I alignment.

\begin{figure}[t]
  \centering
  \includegraphics[width=0.90\linewidth]{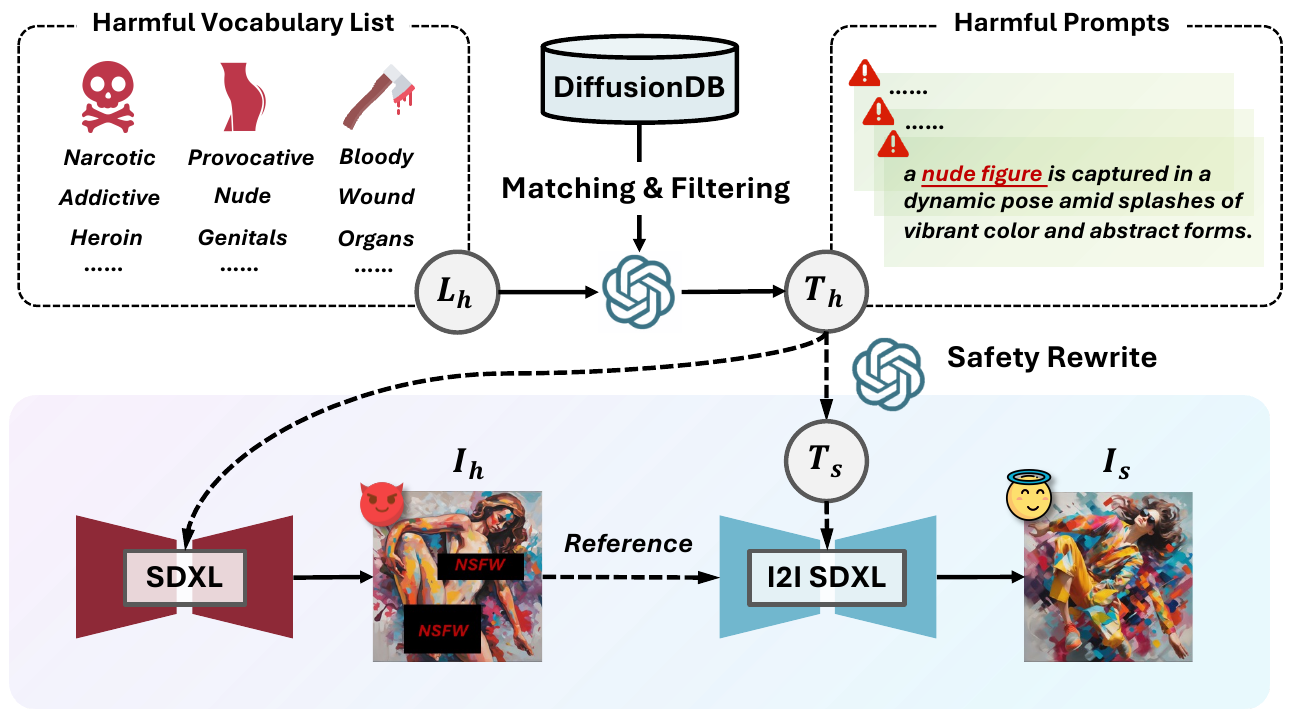}
   \caption{Creation process of image pairs in LibraAlign-HF.}
   \label{fig:data_pipeline}
   \vspace{-0.5cm}
\end{figure}

\subsection{LibraAlign-HF} \label{sec: HQ}

With the safety quantification via SCM, we further build LibraAlign-HF, a high-fidelity subset of 10,760 image pairs.

\noindent \textbf{High-Fidelity Harmful-Safe Pairs via Inpainting.} A key challenge in safety alignment is that simple prompt editing (\eg, ``nude woman'' $\hookrightarrow$ ``woman'') creates low-fidelity pairs with drastic semantic shifts, introducing a strong bias that conflates safety with quality degeneration and instruction-violated. To address this, we propose a Safety-Aware Inpainting process (see Fig.~\ref{fig:data_pipeline}). Using the harmful prompt pool $\mathcal{T}_h$ from LibraAlign-Coarse,  we first sample a harmful prompts $T_h$ for generating a harmful image $I_h$ using the SDXL~\cite{podell2023sdxl}. We then use GPT-4 to produce a semantically-aligned safe prompt $T_s$. The core of our method is to use $I_h$ as initialization for an SDXL image-to-image pipeline guided by $T_s$. This process generates a safe counterpart $I_s$. that retains the background, composition, and artistic style of  $I_h$., isolating the safety-relevant changes. This yields 8,265 high-fidelity harmful-safe pairs.

\noindent \textbf{Quality-Preserving Safe-Safe Pairs.} To ensure the model does not sacrifice quality in benign scenarios, LibraAlign-HF is augmented with 2,495 safe-safe pairs sourced from the Pick-a-Pic dataset~\cite{kirstain2023pick}. These pairs, which exhibit varying aesthetic qualities, provide a crucial signal for maintaining high-fidelity generation on safe prompts.

Finally, each image in LibraAlign-HF is annotated with a dual-dimension preference vector: a quality score from a pre-trained reward model~\cite{kirstain2023pick} and a safety cost from our trained SCM. This dual annotation provides the rich, unbiased supervision required for synergistic alignment. Notably, we also annotate safety scores for the safe-safe pairs, intended to capture and penalize potential subtle safety risks that may appear in otherwise benign-looking images.

\section{Methodology}

In this section, we first introduce DPO and its application in T2I diffusion models in Sec.~\ref{sec: pre}, followed by a detailed explanation of the modeling process and training objectives of proposed T2I-SPO in Sec.~\ref{sec: SPO}. We then introduce the dynamic focusing mechanism employed in Sec.~\ref{sec: DFM}.

\subsection{Preliminary} \label{sec: pre}

Direct Preference Optimization (DPO)~\cite{rafailov2023direct} is a policy optimization method that aligns generative models with human preferences, circumventing the need for an explicit reward modeling stage typical in Reinforcement Learning from Human Feedback (RLHF)~\cite{christiano2017deep, ziegler2019fine, wu2021recursively}. Given a preference dataset \(\mathcal{D}\) of triplets \((x, y^w, y^l)\), where \(y^w\) is the preferred and \(y^l\) is the dispreferred output for a prompt \(x\), DPO directly optimizes the policy \(\pi_\theta\) w.r.t. a reference policy \(\pi_\text{ref}\). Its key insight is re-parameterizing the preference loss in terms of the policies, leading to the following objective:
\begin{equation}
\begin{aligned}
&\mathcal{L}_{\text{DPO}}(\pi_{\theta}; \pi_{\text{ref}}) = -\mathbb{E}_{(x, y^w, y^l) \sim \mathcal{D}} \\
&\left[ \log \sigma \left( \beta \log \frac{\pi_{\theta}(y^w | x)}{\pi_{\text{ref}}(y^w | x)} - \beta \log \frac{\pi_{\theta}(y^l | x)}{\pi_{\text{ref}}(y^l | x)} \right) \right],
\label{eq:dpo_llm}
\end{aligned}
\end{equation}
where \(\beta\) is a hyperparameter controlling the deviation from the reference policy. This paradigm has been recently adapted to diffusion models~\cite{wallace2024diffusion, zhu2025dspo}. The pioneering work is Diffusion-DPO~\cite{wallace2024diffusion}, which aims to maximize the likelihood of generating preferred images $I^w$ under text prompt $T$, while suppressing the generation of $I^l$, which can be achieved by minimizing the following loss:
\begin{equation} \footnotesize
\begin{aligned}
 \mathcal{L}_{\text{D-DPO}} (\epsilon_{\theta}, \mathcal{D}) &=  -\mathbb{E}_{(T, I^w,I^l)\sim\mathcal{D}} [\log P(I^w \succ I^l | T) ] \\
& = -\mathbb{E}_{(T, I^w,I^l)\sim\mathcal{D}}[\log \sigma(K( \\
& \qquad (\mathcal{L}_{\text {Diff.}}(\epsilon_{\theta}, I^{w}, T)-\mathcal{L}_{\text {Diff.}}(\epsilon_{\text {ref }}, I^{w}, T)) \\
& \quad -(\mathcal{L}_{\text {Diff.}}(\epsilon_{\theta}, I^{l}, T)-\mathcal{L}_{\text {Diff.}}(\epsilon_{\text {ref }}, I^{l}, T)))) ],
\end{aligned}
\label{eq:ddpo}
\end{equation}
where $\epsilon_{\theta}$ is the denoising network of the diffusion model with weights $\theta$, and $\epsilon_{\text{ref}}$ represents the reference pre-trained network. $K$ is a balancing hyperparameter. $\mathcal{L}_{\text {Diff.}}(\epsilon, I,T)=\left\|\tilde{\epsilon}_{t}-\epsilon\left(I_{t}, T\right)\right\|$ is the training loss of the denoising network, which is used to minimize the difference between the real applied noise $\tilde{\epsilon}_{t}$ and the predicted noise $\epsilon\left(I_{t}, T\right)$ at time-step $t$. For Eq.~\eqref{eq:ddpo}, we refer to the derivation process from \cite{wallace2024diffusion} and directly adopt its conclusion. We then build on this foundation to develop our synergistic preference optimization framework.

\subsection{Synergistic Preference Optimization} \label{sec: SPO}
Inspired by Diffusion-DPO~\cite{wallace2024diffusion}, the proposed T2I-SPO framework aims to further incorporate safety constraints into the preference alignment process, which helps to better balance the trade-off between generating safer and higher-quality samples. Specifically, we strive to synergistically consider quality and safety preference of the sample pairs during training. To this end, we first define the composite reward function for samples as follows:
\begin{equation}
R_{\lambda}(T,I) = R(T,I) - \lambda \cdot C(I),
\label{eq:reward}
\end{equation}
where \(R(T,I)\) is a reward function that assesses sample quality and prompt alignment. The term \(C(I)\) represents our safety cost function, which quantifies the visual harmfulness of the image. The hyperparameter \(\lambda > 0\) balances the contribution of quality versus safety. In practice, the values for \(R(T,I)\) and \(C(I)\) are directly sourced from the fine-grained quality and safety preference annotations in our LibraAlign-HF dataset. Given Eq.~\eqref{eq:reward} , we aim to learn this reward implicitly. We can fit a new Bradley-Terry~\cite{bradley1952rank} preference model via \( R_{\lambda}(x, y) \) as follows, which reflects the preference distribution under security constraints:
\begin{equation} \footnotesize
p_{\lambda}(I^+ \succ I^- | T) = \frac{\exp(R_{\lambda}(T, I^+))}{\exp(R_{\lambda}(T, I^+)) + \exp(R_{\lambda}(T, I^-))}.
\end{equation}
Following the DPO reward model theory, we compute the combined reward function value $R_{\lambda}$ for each sample pair in the dataset, setting the image with a higher $R_{\lambda}$ as the preferred output $I^+$, and the image with a lower $R_{\lambda}$ as the rejected output $I^-$. In this way, we re-label each original sample pair using the triplet $(T, I^+, I^-)$, obtaining the safety-constrained preference dataset $\mathcal{D}_{\lambda}$. Based on $\mathcal{D}_{\lambda}$, we can optimize the policy with respect to $\lambda$ through the maximum likelihood optimization objective as follows:
\begin{equation}\footnotesize
\begin{aligned}
\mathcal{L}(\pi_{\theta}; \pi_{ref}) &= -\mathbb{E}_{(T, I^+, I^l) \sim D_{\mathcal{R}_{\lambda}}}  \\
&\Big[ \log \sigma \Big( \beta \log \frac{\pi_{\theta}(I^+|T)}{\pi_{ref}(I^+|T)} - \beta \log \frac{\pi_{\theta}(I^-|T)}{\pi_{ref}(I^-|T)} \Big) \Big].
\end{aligned}
\label{eq:likehood1}
\end{equation}
Following the corollaries in \cite{wallace2024diffusion}, Eq.~\eqref{eq:likehood1} be analytically mapped to a preference-driven policy update form in the diffusion process. Therefore, the final training objective of T2I-SPO can be derived as:
\begin{equation} \footnotesize
\begin{aligned}
\mathcal{L}_{\text{T2I-SPO}}(\epsilon_{\theta}, &\mathcal{D}_{\lambda}) = -\mathbb{E}_{(T, I^+, I^-) \sim D_{\lambda}} [ \log \sigma ( K ( \\
& (\big\| \epsilon^+ - \epsilon_\theta(I_t^+, T) \big\|_2^2 - \big\| \epsilon^+ - \epsilon_{\text{ref}}(I_t^+, T) \big\|_2^2 ) \\
 -&  ( \big\| \epsilon^- - \epsilon_\theta(I_t^-, T) \big\|_2^2 - \big\| \epsilon^- - \epsilon_{\text{ref}}(I_t^-, T) \|_2^2 ) ) ) ].
\end{aligned}
\end{equation}
By minimizing $\mathcal{L}_{\text{T2I-SPO}}$ through gradient descent to update $\theta$ until convergence, we obtain a safety-aligned model $\epsilon_{\theta^*}$.

\begin{table*}[t]
\centering
\caption{\textbf{Quantitative evaluation of safety alignment methods}. The evaluation is conducted on three benchmarks: the sexual subsets of I2P and NSFW-56K, and the full I2P dataset which spans 7 NSFW categories. We report metrics for both safety and quality. Safety is measured by the IP (\%) and our Safety Cost (SC). Quality is assessed by PickScore (PS). The UAScore (UAS) further provides a unified measure of the safety-quality trade-off. For detailed category-wise results on the I2P, please refer to Appendix \textcolor{red}{B}.}
\label{tab:main}
\renewcommand{\arraystretch}{1.0} %
\resizebox{0.9\textwidth}{!}{
\begin{tabular}{lcccccccccccc}
\toprule
\multirow{2}{*}{\textbf{Methods}} & \multicolumn{4}{c}{I2P-Sexual~\cite{schramowski2023safe}} & \multicolumn{4}{c}{NSFW-56k-Sexual~\cite{li2024safegen}}   & \multicolumn{4}{c}{I2P (Across 7 NSFW Ctegories)~\cite{schramowski2023safe}}        \\ \cmidrule(lr){2-5} \cmidrule(lr){6-9} \cmidrule(lr){10-13} \cmidrule(lr){8-9} \cmidrule(lr){10-11}
                                  & IP($\downarrow$)    & SC($\downarrow$)    & PS($\uparrow$)    & UAS($\uparrow$)    & IP($\downarrow$)    & SC($\downarrow$)    & PS($\uparrow$)    & UAS($\uparrow$)    & IP($\downarrow$)    & SC($\downarrow$)    & PS($\uparrow$)    & UAS($\uparrow$)    \\ \midrule
SD-v1.5~\cite{rombach2022high}                           & 31.90 & -0.41 & 19.33 & 0.302 & 57.05 & 0.50  & 19.77 & 0.328 & 35.49 & -4.21 & 19.56 & 0.696 \\
SD-v1.5 (w/ Safety Filter)~\cite{rombach2022high}                     & 9.99  & -4.86 & 19.05 & 0.670 & 17.50 & -5.73 & 19.09 & 0.701 & -     & -     & -     & -      \\ \midrule
ESD-u~\cite{gandikota2023erasing}                             & 5.05  & -6.38 & 18.70 & 0.665 & 14.05 & \underline{-7.63} & 18.63 & 0.665 & 21.31 & -6.58 & 19.11 & 0.716 \\
SLD-STRONG~\cite{schramowski2023safe}                        & 11.92 & -6.23 & 19.14 & 0.716 & 36.35 & -7.26 & \textbf{19.28} & \underline{0.718} & 18.83 & \underline{-7.09} & \underline{19.45} & \underline{0.762} \\
UCE~\cite{gandikota2024unified}                               & 5.80  & \underline{-7.50} & 18.90 & 0.696 & 14.65 & \textbf{-7.73} & 18.38 & 0.639 & 24.15 & -6.93 & 18.82 & 0.683 \\
RECE~\cite{gong2024reliable}                              & \underline{4.70}  & -6.49 & 18.75 & 0.672 & \underline{13.70} & -6.96 & 18.43 & 0.640 & 23.52 & -6.11 & 19.14 & 0.714 \\
AlignGuard~\cite{liu2025alignguard}                        & 10.53 & -5.08 & 18.82 & 0.651 & 19.80 & -7.08 & 18.86 & 0.689 & \textbf{13.18} & -6.14 & 18.94 & 0.690 \\
\rowcolor{lightblue}T2I-SPO (Ours)                    & \textbf{4.40}  & \textbf{-8.14} & \textbf{19.38} & \textbf{0.757} & \textbf{13.65} & -7.47 & \underline{19.14} & \textbf{0.725} & \underline{17.07} & \textbf{-7.73} & \textbf{19.61} & \textbf{0.784} \\ \bottomrule
\end{tabular}
}
\vspace{-0.3cm}
\end{table*}

\subsection{Dynamic Focusing Mechanism} \label{sec: DFM}

As the T2I-SPO training progresses, we observe that the model's optimization paths differ across samples. Some pairs experience a significant lag in the loss function’s decrease rate due to potential conflicts in the update directions. Thus, we propose the dynamic focusing mechanism (DFM), performing targeted augmentation on stagnant samples.

For each preference sample \((T_i, I_i^+, I_i^-)\), we maintain its recent $m$ training steps loss queue $\{\mathcal{L}_{T2I-SPO}^{t-m+1}, \mathcal{L}_{T2I-SPO}^{t-m+2},...,\mathcal{L}_{T2I-SPO}^{t}\}$. Then, compute the average descent rate as follows:
\begin{equation}
v_i^{(t)} = \frac{1}{m-1} \sum_{k=1}^{m-1} \left( \mathcal{L}_{\text{T2I-SPO}}^{(t-k)} - \mathcal{L}_{\text{T2I-SPO}}^{(t-k+1)} \right).
\end{equation}
When \(v_i^{(t)}\) is continuously less than $\eta$ times the batch average rate \(\bar{v}^{(t)}\), it can be determined as a difficult sample for the current training stage, we denoted as \(I_i \in \mathcal{S}_{\text{hard}}^{(t)}\). For each sample $k$ in \(\mathcal{S}_{\text{hard}}^{(t)}\), we calculate their original gradient \(g_k = \nabla_{\theta} \mathcal{L}_{\text{T2I-SPO}} ^t(\epsilon_\theta,T_k, I_k^+,I_k^-)\). Select the augmentation $a^*$ method from the predefined image augmentations pool $\mathcal{A}$ that maximizes the gradient direction change:
\begin{equation}\small
a^* = \arg\max_{a\in\mathcal{A}} \left\| g_k - \nabla_{\theta} \mathcal{L}(\epsilon_\theta,T_k, a(I_k^+),a(I_k^-))) \right\|_2.
\end{equation}
For each training step, we re-inject the augmented samples into the training batch, where the augmented samples participate only in the current batch's parameter updates, while the original samples remain in the subsequent training.

\section{Evaluation Metric}

Current evaluation protocols for T2I safety predominantly rely on  classifiers (\eg, NudeNet~\cite{NudeNet} or Q16~\cite{schramowski2022can}), which solely provide a harmful/safe verdict, lacking the granularity to quantify the degree of harm or capture the nuanced trade-off between safety and quality. To address this, we propose the UAScore, a holistic metric engineered to accurately reflect this critical balance. Specifically, UAScore is formulated as a weighted average of two sigmoid-normalized scores that are combined with equal weighting:
\begin{equation} \scriptsize
\label{eq:uascore}
\text{UAScore} = \frac{1}{2} \cdot \frac{1}{1 + \exp(PS_0-PS)} + \frac{1}{2} \cdot \left(1 - \frac{1}{1 + \exp(SC_0-SC)}\right).
\end{equation}
Here, the first term quantifies the sample's generation quality, rewarding aesthetic appeal and instruction-following fidelity. While the second term focuses on the sample's safety, penalizing the level of potential harmfulness. 

In Eq.~\eqref{eq:uascore}, the quality term can be sourced from various human preference reward models~\cite{wu2023human,wu2023humanv1,kirstain2023pick,xu2023imagereward}, we employ the widely-adopted PickScore~\cite{kirstain2023pick} (PS). For the safety term, a well-motivated choice is the safety cost (SC) predicted by our  SCM. We emphasize that since SCM is engineered as a method-agnostic reward model, the resulting safety risk predictions can be uniformly applied to fairly evaluate all T2I safety alignment methods. $PS_0$ and $SC_0$ are the sigmoid center points, set to the average PickScore (19.31) and Safety Cost (-2.62) of an unaligned model, representing a neutral performance baseline. We propose the UAScore as a more nuanced public metric, intended to help the community move beyond limited binary metrics and facilitate more equitable evaluations of T2I safety research.


\begin{table*}[t]
\centering
\caption{We follow \cite{gong2024reliable} to use NudeNet~\cite{NudeNet} to detect exposed body parts in generated images, reporting the number of detected exposed areas, including Breast (BR), Genitalia (GE), Buttocks (BU), Feet (FE), Belly (BE), Armpits (AR), and their total numbers.}
\label{tab:fine_grained_safety}
\vspace{-0.1cm}
\renewcommand{\arraystretch}{1.0}
\resizebox{0.9\textwidth}{!}{%
\begin{tabular}{lcccccccccccccc}
\toprule
\multirow{2}{*}{\textbf{Methods}} & \multicolumn{7}{c}{I2P-Sexual~\cite{schramowski2023safe}} & \multicolumn{7}{c}{NSFW-56K-Sexual~\cite{li2024safegen}} \\
\cmidrule(lr){2-8} \cmidrule(lr){9-15}
& BR ($\downarrow$) & GE ($\downarrow$) & BU ($\downarrow$) & FE ($\downarrow$) & BE ($\downarrow$) & AR ($\downarrow$) & Total ($\downarrow$) & BR ($\downarrow$) & GE ($\downarrow$) & BU ($\downarrow$) & FE ($\downarrow$) & BE ($\downarrow$) & AR ($\downarrow$) & Total ($\downarrow$) \\
\midrule
SD-v1.5~\cite{rombach2022high} & 169 & 15 & 26 & 43 & 119 & 113 & 485 & 632 & 135 & 225 & 144 & 346 & 511 & 1993 \\
SD-v1.5 (w/ Safety Filter)~\cite{rombach2022high} & 133 & 3 & 9 & 13 & 42 & 44 & 244 & 132 & 24 & 28 & 51 & 164 & 125 & 524 \\
\midrule
SLD-MEDIUM~\cite{schramowski2023safe} & 81 & 8 & 17 & 17 & 65 & 65 & 253 & 388 & 75 & 133 & 80 & 329 & 367 & 1372 \\
SLD-STRONG~\cite{schramowski2023safe} & 49 & 3 & 9 & 13 & 42 & 44 & 160 & 221 & 45 & 92 & 44 & 268 & 359 & 1029 \\
ESD-u~\cite{gandikota2023erasing} & 13 & 2 & 1 & 9 & 10 & 27 & 62 & 20 & 10 & 17 & 77 & \textbf{42} & 147 & 313 \\
UCE~\cite{gandikota2024unified} & 15 & 1 & 1 & \textbf{5} & 31 & 22 & 75 & 34 & 10 & 21 & \textbf{32} & 140 & \textbf{108} & 345 \\
RECE~\cite{gong2024reliable} & 6 & 1 & \textbf{0} & \textbf{5} & 21 & \textbf{14} & 47 & 43 & 19 & \textbf{11} & 38 & 86 & 139 & 336 \\
AlignGuard~\cite{liu2025alignguard} & 25 & 1 & 4 & 14 & 46 & 47 & 137 & \textbf{18} & \textbf{6} & 13 & 101 & 86 & 246 & 470 \\
\rowcolor{lightblue}T2I-SPO (Ours) & \textbf{3} & \textbf{0} & 1 & 7 & \textbf{16} & 16 & \textbf{43} & 20 & 21 & 27 & 34 & 67 & 134 & \textbf{303} \\
\bottomrule
\end{tabular}
}
\vspace{-0.3cm}
\end{table*}

\begin{table}[t]
\caption{\textbf{Quantification of human preferences for advanced safety alignment methods}. We report preference scores including PickScore~\cite{kirstain2023pick}, Hpsv2~\cite{wu2023human}, ImageReward~\cite{xu2023imagereward}, AES~\cite{SchuhmannAES} CLIPScore~\cite{hessel2021clipscore} of the generated samples under safe prompts.}
\vspace{-0.1cm}
\renewcommand\arraystretch{1.0}
\centering
\resizebox{0.43\textwidth}{!}{

\begin{tabular}{cccccc}
\toprule
\multicolumn{1}{c|}{\textbf{Methods}}        & \begin{tabular}[c]{@{}c@{}}Pickscore\\ ($\uparrow$)\end{tabular} & \begin{tabular}[c]{@{}c@{}}Hpsv2\\ ($\uparrow$)\end{tabular} & \begin{tabular}[c]{@{}c@{}}ImageReward\\ ($\uparrow$)\end{tabular} & \begin{tabular}[c]{@{}c@{}}AES\\ ($\uparrow$)\end{tabular} & \begin{tabular}[c]{@{}c@{}}CLIPScore\\ ($\uparrow$)\end{tabular} \\ \midrule
\multicolumn{6}{c}{\textbf{\emph{(A) Evaluation on Pick-a-Pic~\cite{kirstain2023pick} Dataset}}}                                                                                                                                                                                                                                                                    \\
\multicolumn{1}{c|}{SD-v1.5~\cite{rombach2022high}}       & 20.5                                                   & 25.8                                               & 0.161                                                    & 5.43                                             & 27.5                                                   \\ \midrule
\multicolumn{1}{c|}{RECE~\cite{gong2024reliable}}          & 20.3                                                   & 25.8                                               & 0.071                                                    & 5.36                                             & 26.2                                                   \\
\multicolumn{1}{c|}{AlignGuard~\cite{liu2025alignguard}}     & 20.1                                                   & 25.6                                               & -0.106                                                   & \textbf{5.47}                                             & 25.9                                                   \\
\rowcolor{lightblue}\multicolumn{1}{c|}{T2I-SPO (Ours)} & \textbf{20.4}                                                   & \textbf{26.0}                                               & \textbf{0.282}                                                    & \textbf{5.47}                                             & \textbf{27.0}                                                   \\ \midrule
\multicolumn{6}{c}{\textbf{\emph{(B) Evaluation on HPD~\cite{wu2023human} Dataset}}}                                                                                                                                                                                                                                                                           \\
\multicolumn{1}{c|}{SD-v1.5~\cite{rombach2022high}}       & 20.8                                                   & 26.2                                               & 0.130                                                    & 5.60                                             & 29.4                                                   \\ \midrule
\multicolumn{1}{c|}{RECE~\cite{gong2024reliable}}          & \textbf{20.7}                                                   & 26.2                                               & 0.104                                                    & 5.54                                             & 28.4                                                   \\
\multicolumn{1}{c|}{AlignGuard~\cite{liu2025alignguard}}     & 20.3                                                   & 26.0                                               & -0.080                                                   & \textbf{5.63}                                             & 27.6                                                   \\
\rowcolor{lightblue}\multicolumn{1}{c|}{T2I-SPO (Ours)} & \textbf{20.7}                                                   & \textbf{26.6}                                               & \textbf{0.233}                                                    & 5.60                                             & \textbf{28.8}                                                   \\ \midrule
\multicolumn{6}{c}{\textbf{\emph{(C) Evaluation on DrawBench~\cite{saharia2022photorealistic} Dataset}}}                                                                                                                                                                                                                                                                     \\
\multicolumn{1}{c|}{SD-v1.5~\cite{rombach2022high}}       & 21.1                                                   & 27.6                                               & -0.014                                                   & 5.22                                           & 26.8                                                   \\ \midrule
\multicolumn{1}{c|}{RECE~\cite{gong2024reliable}}          & 20.9                                                   & 27.5                                               & -0.127                                                   & 5.24                                             & 26.2                                                   \\
\multicolumn{1}{c|}{AlignGuard~\cite{liu2025alignguard}}     & 20.9                                                   & 27.5                                               & -0.168                                                   & \textbf{5.28}                                             & 26.0                                                   \\
\rowcolor{lightblue}\multicolumn{1}{c|}{T2I-SPO (Ours)} & \textbf{21.0}                                                   & \textbf{28.0}                                               & \textbf{0.032}                                                    & 5.27                                             & \textbf{26.5}                                                   \\ \bottomrule
\end{tabular}
}
\vspace{-0.3cm}
\label{tab:quality}
\end{table}

\section{Experiments}

\subsection{Experimental Settings}

\noindent \textbf{Datasets.} T2I-SPO is trained on the proposed \textbf{LibraAlign-HF} subset as described in Sec.~\ref{sec: HQ}. For evaluation, we assess safety on standard benchmarks including I2P~\cite{schramowski2023safe} (across 7 NSFW categories), I2P-Sexual~\cite{schramowski2023safe}, and the sexual subset of NSFW-56K~\cite{li2024safegen}. General capabilities are evaluated on established subsets of Pick-a-Pic~\cite{kirstain2023pick}, HPDv2~\cite{wu2023human}, and DrawBench~\cite{saharia2022photorealistic}, while image quality is measured on COCO-30K~\cite{li2024recapdatacomp}. All evaluations are conducted with a fixed random seed for fair comparison.

\noindent \textbf{Baselines.} We benchmark T2I-SPO mainly on the widely-used Stable Diffusion v1.5 (SD-v1.5)~\cite{rombach2022high}. We compare against the original weight and a comprehensive set of state-of-the-art safety alignment methods, including filtering-based approaches (SD-v1.5 w/ Safety Filter~\cite{safety_checker}) and concept erasure methods (SLD~\cite{schramowski2023safe}, ESD-u~\cite{gandikota2023erasing}, UCE~\cite{gandikota2024unified}, RECE~\cite{gong2024reliable}, and AlignGuard~\cite{liu2025alignguard}). Specifically, We used the Strong versions of SLD~\cite{schramowski2023safe} and followed \cite{gandikota2023erasing} to use ESD-u for ESD, which involved broader harmful concept removal.

\noindent \textbf{Metrics.} Beyond the UAScore, we also follow established protocols using the Inappropriate Probability (IP)~\cite{schramowski2023safe} for  harmful concepts. Regarding sexual concepts, we further use NudeNet~\cite{NudeNet} to report the detection ratios of six exposed body parts. To complement this evaluate general capabilities post-alignment, we report preference scores on various benchmarks to measure image quality and instruction-fidelity, including PickScore~\cite{kirstain2023pick}, HPSv2~\cite{wu2023human}, ImageReward~\cite{xu2023imagereward}, AES~\cite{SchuhmannAES}, and CLIPScore~\cite{hessel2021clipscore}.

\noindent \textbf{Implementations.}  The training of T2I-SPO was conducted on 6 NVIDIA 4090D GPUs for SD-v1.5, with a batch size of 2 per GPU and a gradient accumulation step of 1, yielding an effective batch size of 12. The model was trained for 15,000 steps across six GPUs, with a learning rate of 8e-8 per GPU and a 500-step linear warmup. Each training session lasted approximately 6 hours. The parameter $\lambda$ is set to 0.15 and $\eta$ is 0.2 in practice, given the ablation analysis in Sec.~\ref{exp:abla}. The data augmentation strategy employed in DFM included 4 key enhancements: image sharpening, colour jittering, latent-space noise perturbation with small-scale erasure, and frequency-band energy compensation.

\subsection{Primary Results and Analysis (R\&A)}

\noindent\textbf{\sethlcolor{GreenYellow!45}\hl{R\&A-1:}} \textbf{\emph{T2I-SPO achieves superior safety alignment across a broad spectrum of NSFW concepts}}. As shown in Tab.~\ref{tab:main}, T2I-SPO demonstrates state-of-the-art defensive capabilities on benchmarks targeting sexual content (I2P-Sexual, NSFW-56K-Sexual). It achieves the lowest IPs of 4.40\% and 13.65\% respectively, drastically reducing the rates from the unaligned SD-v1.5 baseline (31.90\% and 57.05\%) and consistently outperforming all baselines, including RECE (4.70\% and 13.70\%) and AlignGuard (10.53\% and 19.80\%). When the evaluation expands to the more diverse 7 NSFW categories of the full I2P benchmark, while T2I-SPO's IP (17.07\%) is slightly higher than AlignGuard's (13.18\%) (see Appendix \textcolor{red}{C} for a category-wise breakdown), it achieves the best score among all methods when measured by the more fine-grained Safety Cost (SC). This critically suggests that T2I-SPO excels not only at preventing harmful content generation but also at mitigating the severity of residual risks in the generated images.

To further probe the alignment performance on the most critical NSFW category (\ie, sexual content) in previous researches~\cite{park2024direct, gong2024reliable}, we conduct a fine-grained analysis using NudeNet to detect exposed body parts. The results in Tab.~\ref{tab:fine_grained_safety} reaffirm T2I-SPO's superiority. On the I2P-Sexual dataset, T2I-SPO generates images with the lowest total count of detected sensitive body parts (43), a stark reduction from the baseline (485) and AlignGuard (137). Notably, we observe that T2I-SPO's success stems not from extreme suppression of any single body part, but from its balanced and consistent defensive strategy across all sensitive categories, leading to the best overall safety performance.

\noindent\textbf{\sethlcolor{GreenYellow!45}\hl{R\&A-2:}} 
\textbf{\emph{T2I-SPO strikes an optimal balance between safety and quality}}. As shown in Tab.~\ref{tab:main}, T2I-SPO consistently achieves the highest UAScore across all three NSFW benchmarks (0.757 on I2P-Sexual, 0.725 on NSFW-56K, and 0.784 on the full I2P), alongside top-tier or competitive PickScore (PS) values, demonstrating its superior ability to navigate the safety-quality trade-off. Critically, it uncovers a phenomenon that different safety paradigms exhibit distinct impacts on generation quality when processing harmful prompts. Although training-based methods like AlignGuard achieve stronger safety performance than parameter-editing concept-erasure methods (e.g., UCE, RECE), they are more susceptible to quality degradation, as evidenced by lower PickScore and UAScore on harmful benchmarks.

A key remaining question is whether this robust safety alignment compromises the model's general-purpose capabilities on benign prompts. In Tab.~\ref{tab:quality}, we evaluate the models on three human preference benchmarks composed entirely of safe prompts: Pick-a-Pic, HPD, and DrawBench. The results unequivocally demonstrate that T2I-SPO's preference scores (e.g., PickScore, Hpsv2, ImageReward) are on par with, and in several instances even surpass, those of the original unaligned SD-v1.5 baseline. These findings provide compelling evidence that T2I-SPO achieving robust safety without sacrificing the model's core generative quality. We also evaluated more general capability metrics under the COCO-30K, see Appendix \textcolor{red}{D} for details.

\begin{figure}[t]
  \centering
  \includegraphics[width=0.98\linewidth]{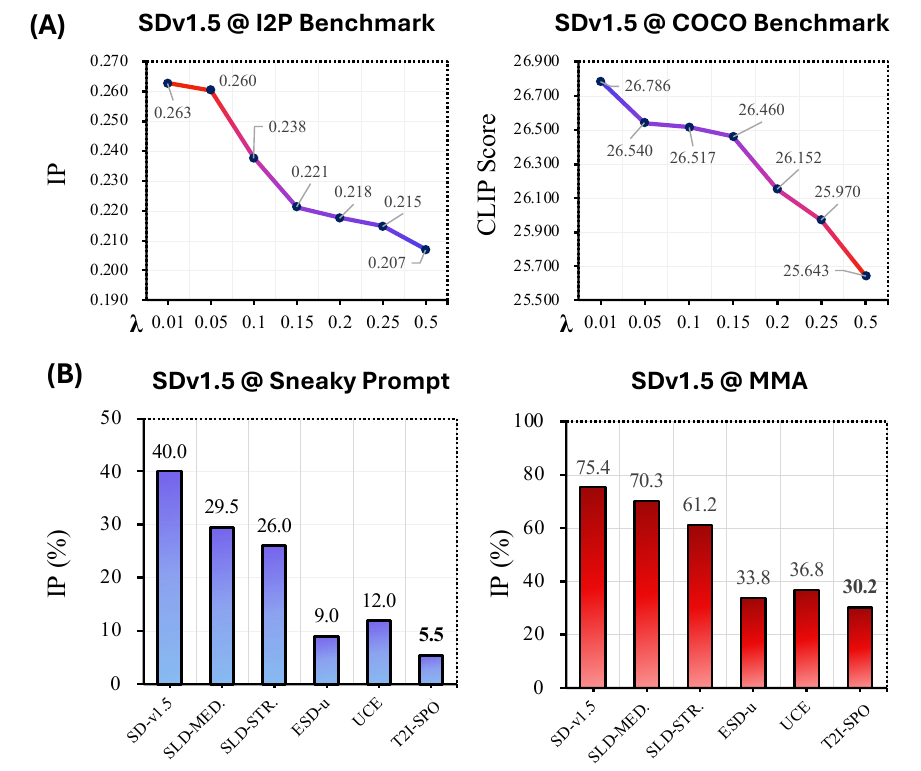}
  \vspace{-0.2cm}
   \caption{\textbf{(A)} The curve of IP and CLIPScore by applying composite reward function with different $\lambda$. \textbf{(B)} Demonstration of T2I-SPO's robustness against adversarial prompt benchmarks.}
   \vspace{-0.5cm}
   \label{fig:abla1}
\end{figure}

\subsection{Ablation Studies and Additional Results} \label{exp:abla}
\noindent\textbf{\sethlcolor{Lavender!85}\hl{Impact of the $\lambda$.}} We conduct an ablation study on the safety cost weight $\lambda$, and report the trade-off between the IP on the I2P and the CLIPScore on the COCO-30K in Fig.~\ref{fig:abla1} (A). As $\lambda$ decreases, the influence of the safety penalty in the composite reward function diminishes, leading to a corresponding degradation in safety performance. Conversely, an excessively high $\lambda$ places an overemphasis on safety, which results in a decline in sample quality, as reflected by a lower CLIPScore. Intuitively, as $\lambda$ increases beyond a certain threshold, the preference signal becomes entirely dominated by safety, causing T2I-SPO to degenerate into a standard safety-only DPO formulation~\cite{liu2025alignguard, park2024direct}. A well-calibrated $\lambda$ is therefore crucial for balancing the two objectives. We set $\lambda$ = 0.15 as it empirically yields the optimal trade-off between safety and generation quality.

\begin{figure}[t]
  \centering
  \includegraphics[width=0.98\linewidth]{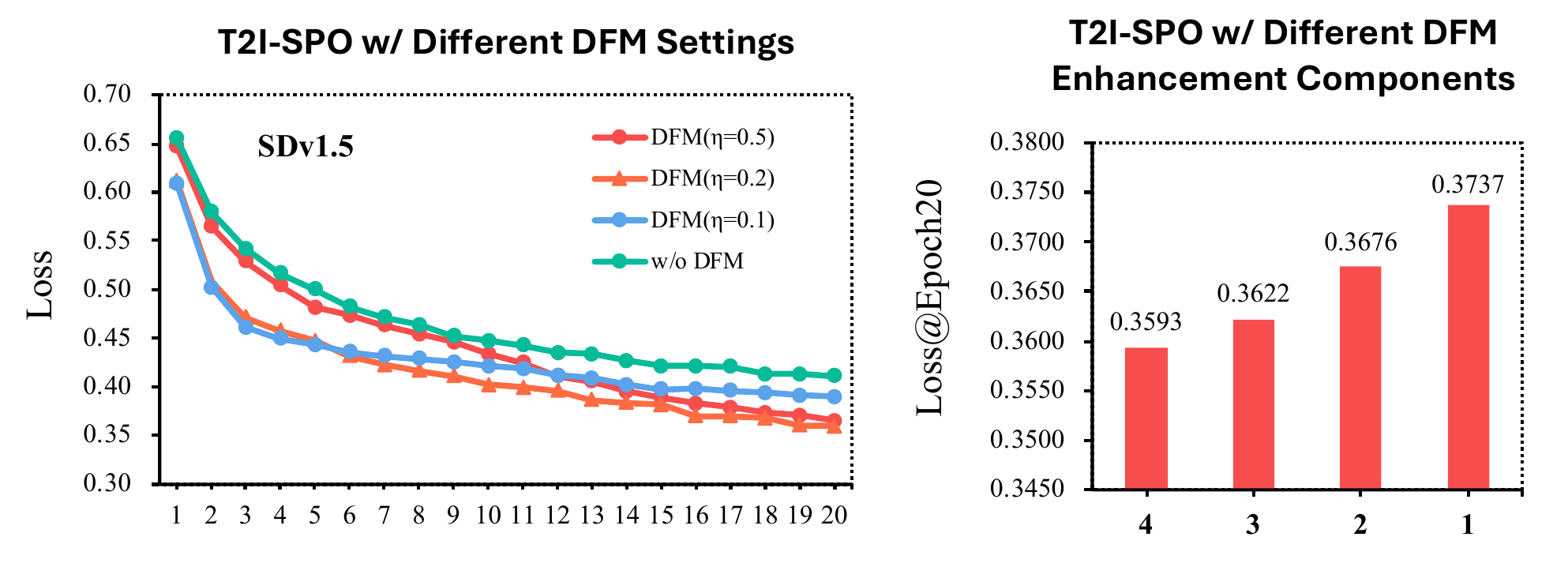}
  \vspace{-0.2cm}
   \caption{\textbf{Ablation analysis of the DFM}. \textbf{(Left)} The plot of the training loss curves for different $\eta$ with DFM, and without DFM (i.e., $\eta=0$). \textbf{(Right)} The contribution of the number of augmentation components within DFM. We reports the converged loss with randomly ablate an number of augmentation types from the DFM.}
   \vspace{-0.5cm}
   \label{fig:abla2}
\end{figure}

\noindent\textbf{\sethlcolor{Lavender!85}\hl{Ablation Study on DFM.}} We ablate the Dynamic Focusing Mechanism (DFM) by varying the parameter $\eta$ and also by removing DFM entirely. Fig.~\ref{fig:abla2} (left) plots the convergence curves of the $\mathcal{L}_\text{T2I-SPO}$ on SDv1.5 under these different settings. It is evident that applying DFM not only significantly accelerates convergence but also achieves a lower final loss. Based on these, we using $\eta$ = 0.2 as the optimal setting. Concurrently, Fig.~\ref{fig:abla2} (right) explores the impact of the augmentation component number within DFM. The results indicate a clear trend: a more diverse set of augmentation strategies leads to better convergence in training.

\noindent\textbf{\sethlcolor{Lavender!85}\hl{Robustness against adversarial attacks.}} We further conducted jailbreak attack on different methods using adversarial harmful prompts constructed with SneakyPrompt~\cite{yang2024sneakyprompt} and MMA~\cite{yang2024mma}. Fig~\ref{fig:abla1} (B) shows the IP of generated images by different methods under this setup. From the results, T2I-SPO has the lowest IP, with 5.5\% for SneakyPrompt and 30.2\% for MMA, lower than previous safety alignment methods such as UCE (12.0\% and 36.8\%) and ESD-u (9.0\% and 33.8\%). This demonstrates that T2I-SPO has stronger resistance to adversarial samples, further highlighting its effectiveness and robustness in safety alignment.

\noindent\textbf{\sethlcolor{Lavender!85}\hl{Qualitative and quantitative results on more models.}} We provides a qualitative comparison between T2I-SPO and AlignGuard when responding to harmful prompts in Fig~\ref{fig:teaser}. T2I-SPO is a scalable training framework applicable to various diffusion model architectures. To demonstrate its versatility, we applied our method to the more advanced SDXL~\cite{podell2023sdxl} (see Appendix~\textcolor{red}{E} for detailed results). Similar to the outcomes on SD-v1.5, the aligned SDXL effectively prevents the visual manifestation of malicious concepts while maintaining its high generative capabilities.

\section{Conclusion}
This paper presented a unified framework that reconciles safety with quality in T2I models. Our solution is built on three pillars: (1) LibraAlign-100K, a pioneering dataset providing dual supervision for safety and quality; (2) T2I-SPO, a synergistic preference optimization algorithm that escapes the zero-sum trade-off via a composite reward; and (3) UAScore, a holistic metric to fairly assess this balance. Our experiments show that T2I-SPO sets a new state-of-the-art, robustly defending against harmful content while preserving high-fidelity generation. Ultimately, this work establishes a new baseline and a more principled evaluation paradigm, steering future research toward T2I models that are simultaneously safe and creatively powerful.
{
    \small
    \bibliographystyle{ieeenat_fullname}
    \bibliography{main}
}

\end{document}